\documentclass{article}
\usepackage{spconf,amsmath,graphicx,hyperref,amsfonts}
\usepackage{booktabs,multirow}
\usepackage{graphicx}
\usepackage{makecell}
\usepackage{float}

\usepackage[table]{xcolor} 
\definecolor{WinGreen}{RGB}{0,180,0}
\definecolor{RunGreen}{RGB}{200,235,200}
\newcommand{\win}[1]{\cellcolor{WinGreen}\textbf{#1}}
\newcommand{\ru}[1]{\cellcolor{RunGreen}#1}

\title{Constrained Paraphrase Consistency for LLM Hallucination Detection}
\name{
Shanshan Lin$^{1,\star}$ \quad Dongsheng Hong$^{1,\star}$ \quad Sibo Ju$^{1}$ \quad Chao Chen$^{2}$ 
\quad Xi Zhang$^{3}$
\quad Xiangwen Liao$^{1,\dagger}$
}

\address{
        $^{1}$Fuzhou University \qquad 
        $^{2}$Harbin Institute of Technology (Shenzhen)
        \\
        $^{3}$Beijing University of Posts and Telecommunications
        \\
        $^{\star}$Equally contribution \qquad 
        $^{\dagger}$Corresponding author
        }
      
\begin{document}
%
\maketitle
\begin{abstract}
Large language models (LLMs) can generate factually inconsistent claims, motivating accurate and scalable hallucination detectors. 
Prior work largely enlarges training sets via synthesis or new annotations, introducing increasing cost and potential bias while underusing the consistency implied by semantically equivalent paraphrases. 
We propose \emph{Consistency-Constrained Hallucination Detector} (CCHD), which formulates training as a constrained optimization problem. 
The standard cross-entropy on original document-claim pairs is complemented by 
(i) \emph{paraphrase-consistency} constraints bounding divergence across paraphrased views,
and (ii) \emph{label-preservation} constraints tying paraphrases to ground truth. 
We solve the problem by gradient descent-ascent over model parameters and per-view Lagrange multipliers, adding only a few scalar dual variables and no inference-time overhead. 
With DeBERTa and Flan-T5 backbones, CCHD consistently outperforms strong baselines (FactCG, MiniCheck, and AlignScore) on standard factuality benchmarks, demonstrating its superiority on hallucination detection. 


\end{abstract}


\section{Introduction}
\label{sec:intro}
\begin{figure}
    \centering
    \includegraphics[width=0.95\linewidth]{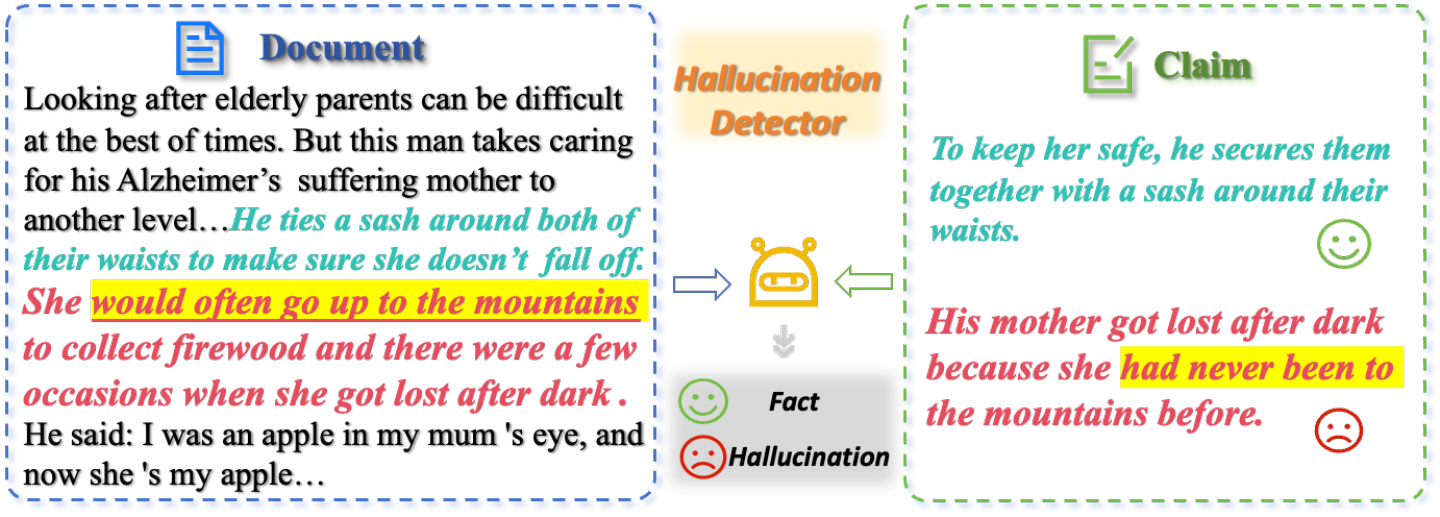}
    \caption{An example from the hallucination detection task.}
    \label{fig:example}
\end{figure}

Large language models (LLMs) generate fluent text for tasks such as abstractive summarization \cite{shakil2024summary_review}. 
However, they are prone to \emph{factual hallucinations}, where outputs appear grammatical but contain unsupported or incorrect information \cite{maynez2020faithfulness}. 
Prior work~\cite{kryscinski2019factcc,laban2023summedits} reports that up to 30\% of generated summaries contain factual inconsistencies, motivating scalable methods to \emph{automatically detect factual consistency}. 
Fig. \ref{fig:example} shows an example from the hallucination detection task. 

Research on factuality detection follows two paradigms. 
One line prompts powerful LLMs to act as judges via sampling, self-consistency, or critique~\cite{manakul2023selfcheckgpt,muhammed2025selfcheckagent,yang2025metaqa}. 
However, these evaluators are \textit{computationally expensive} because they require multiple LLM calls per instance.
%
Another line of studies train lightweight \emph{discriminative} detectors on document-claim pairs, e.g., NLI-based SummaC~\cite{laban2022summac}, MiniCheck~\cite{tang2024minicheck}, and graph-augmented FactCG~\cite{lei2025factcg}. 
%
This line of work prioritizes \emph{data generation} (via synthetic augmentation or new annotations) to expand coverage, incurring extra costs. 
Moreover, they (i) underexplore structure present in existing data that semantically equivalent paraphrases of a claim should yield consistent predictions for a fixed document; 
and (ii) often treat original and synthetic samples identically, despite potential bias and noise in these synthetic samples.

We address these limitations by introducing \emph{Consistency-Constrained Hallucination Detector} (CCHD).
Our premise is simple: if a document supports (or contradicts) a claim, any semantically equivalent paraphrase should receive the same prediction. 
CCHD formalizes this premise and casts training as a \emph{constrained optimization} problem that enforces prediction-level agreements across multiple paraphrase views.
Concretely, 
CCHD sets the standard cross-entropy on original examples as primary objective, 
while two families of \emph{soft constraints} regulate paraphrased views:
(i) \emph{paraphrase-consistency} constraints that bound the divergence between the original prediction and its paraphrased counterparts; 
and (ii) \emph{label-preservation} constraints that keep each paraphrase aligned with the ground-truth label, with controllable slack to accommodate potential noise in paraphrased text. 
CCHD solves the problem via gradient descent-ascent (GDA) over model parameters and per-view Lagrange multipliers~\cite{lin2020gda,chen2021snx}, adaptively balancing supervision and consistency \textit{without} introducing inference-time overhead. 

Empirically, CCHD outperforms strong baselines, including FactCG \cite{lei2025factcg}, MiniCheck \cite{tang2024minicheck}, and Alignscore \cite{zha2023alignscore} across 11 factuality tasks in the \textit{LLM-AggreFact} benchmark. 
Sensitivity analysis and ablation studies further confirm the importance of each component.

In summary, our contributions are as follows:
(1) We cast hallucination detection as a \emph{constrained optimization} problem that enforces prediction-level agreement across semantically equivalent paraphrases.
(2) We propose CCHD, which instantiates paraphrase views by \emph{back-translation} without extra annotation, and solves the constrained problem with GDA over per-view Lagrange multipliers, adding no inference-time overhead.
(3) With Flan-T5 and DeBERTa backbones, CCHD and its variants consistently outperform MiniCheck, FactCG, and AlignScore on standard factuality benchmarks.

\section{Method}
\label{sec:method}

\begin{figure}
    \centering
    \includegraphics[width=0.95\linewidth]{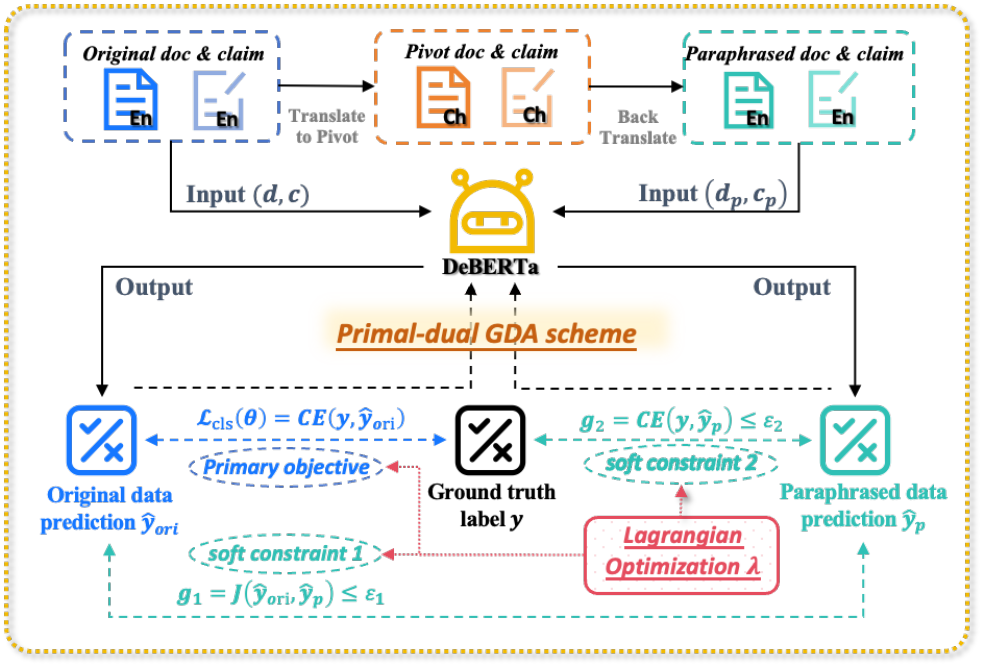}
    \caption{The framework of CCHD.}
    \label{fig:framework}
\end{figure}

Fig.~\ref{fig:framework} shows the key of CCHD framework: 
(i) a constrained training objective that enforces agreement across semantically equivalent paraphrase views (Sec.~\ref{sec:constrained_opt});
and (ii) a Lagrangian optimization scheme to solve the constrained problem (Sec.~\ref{sec:lagrangian_opt_gda}).
Finally, we introduce a practical instantiation of view generation via back-translation in Sec.~\ref{sec:back_translation}.

\subsection{Constrained Training Objective}
\label{sec:constrained_opt}

\textbf{Problem setup.}
Given a document $d$ and an LLM-generated claim $c$, the task is to predict whether $c$ is \emph{supported} or \emph{contradicted} by $d$.
Let $y\in\{0,1\}$ be the ground-truth label and $f_\theta(d,c)\in[0,1]$ the detector's predicted probability that $c$ is consistent with $d$. 
We denote predictions $\hat{y} = f_\theta(d,c)$ for the original input, and
$\hat{y}_p = f_\theta(d_p,c_p)$ for paraphrased $(d_p,c_p)$ via a paraphraser $p\in\mathcal{P}$ (e.g., back-translation~\cite{kryscinski2019factcc}). 


\textbf{Primary objective.}
The primary goal is to minimize the cross-entropy on the original input:
\begin{equation}
\mathcal{L}_{\text{cls}}(\theta) \;=\; \mathrm{CE} \big(y,\; \hat{y}\big),    
\end{equation}
which encourages agreement between $\hat{y}$ and $y$.

\textbf{Consistency constraints.}
Rather than simply appending paraphrased samples to the primary loss~\cite{kryscinski2019factcc,tang2024minicheck}, 
we (i) explicitly encode the \emph{semantic equivalence} between original and paraphrased views, 
and (ii) accommodate potential noise in paraphrases. 
To this end, we impose prediction-level \emph{soft constraints} for each view $p$:
\begin{equation}
\label{eq:c1}
\text{(C1)}\quad 
g_{p,1}(\theta) \; :=\; 
\mathrm{J} \big(\hat{y},\,\hat{y}_{p}\big) - \epsilon_{p,1} \;\le\; 0,
\end{equation}
\begin{equation}
\label{eq:c2}
\text{(C2)}\quad 
g_{p,2}(\theta) \; :=\; 
\mathrm{CE} \big(y,\,\hat{y}_{p}\big) - \epsilon_{p,2} \;\le\; 0,
\end{equation}
where $\mathrm{J}(\mu,\nu)=\tfrac{1}{2}\big(\mathrm{KL}(\mu\|\nu)+\mathrm{KL}(\nu\|\mu)\big)$ is the Jeffreys divergence~\cite{bousquet2023jeffreys}, 
and $\epsilon_{p,1},\epsilon_{p,2} \geq 0$ are tolerances. 

Constraint~\eqref{eq:c1} encourages \emph{paraphrases consistency} by bounding the divergence between prediction distributions before and after paraphrasing. 
Constraint~\eqref{eq:c2} enforces \emph{label preservation} for each paraphrase, 
with slack $\epsilon_{p,2}$ to account for occasional paraphrase noise.
C2 reinforces the assumption that paraphrase does not change the veracity of the claim. 

All constraints are defined at the \textit{prediction} level. 
However, an embedding-level variant (e.g., bounding distance between representations) is also possible without altering the framework, and is experimentally explored in Sec. \ref{sec:sensitivity_analysis}.

\subsection{Lagrangian Optimization via Gradient Descent–Ascent}
\label{sec:lagrangian_opt_gda}

Training under the constraints in Eqs.~\eqref{eq:c1}-\eqref{eq:c2} is posed as a constrained optimization problem: 
\begin{equation}
\min_{\theta}\;\mathcal{L}_{\text{cls}}(\theta)
\quad \text{s.t.} \quad
g_{p,i}(\theta)\le 0
\;\;\;\;\forall\, p\!\in\!\mathcal{P},\; i\!\in\!\{1,2\}.
\end{equation}
To solve it, we form the Lagrangian by introducing nonnegative multipliers
$\boldsymbol{\lambda}=\{\lambda_{p,i}\geq0 \;|\; \forall p,i\}$:
\begin{equation}
\label{eq:lagrangian_obj}
\mathcal{L}(\theta,\boldsymbol{\lambda})
\;=\;
\mathcal{L}_{\text{cls}}(\theta)
\;+\;
\sum_{p}\sum_{i}
\lambda_{p,i}\, g_{p,i}(\theta).
\end{equation}
It introduces only a handful of additional scalar variables (one per constraint), leaving the model architecture unchanged.


The goal is to find a saddle point $(\theta^\star,\boldsymbol{\lambda}^\star)$ that minimizes $\mathcal{L}$ (Eq. \eqref{eq:lagrangian_obj}) in $\theta$ and maximizes it in $\boldsymbol{\lambda}$. 
We adopt a primal-dual \emph{gradient descent-ascent} (GDA) scheme~\cite{lin2020gda}.
In each training iteration,
the updates of $\theta$ and $\boldsymbol{\lambda}$ are defined as:
\begin{equation}
\label{eq:gda_updates}
\begin{aligned}
    \theta &\leftarrow \theta - \eta_0\,\nabla_{\theta}\mathcal{L}(\theta,\boldsymbol{\lambda}),
    \\
    \lambda_{p,i} &\leftarrow \big[\lambda_{p,i} + \eta_{p,i}\, g_{p,i}(\theta)\big]_+,
\end{aligned}
\end{equation}
where $\eta_{0},\eta_{p,i}>0$ are learning rates for the primal and dual variables, respectively. $[\cdot]_+$ denotes projection onto $\mathbb{R}_{\ge 0}$.
This subgradient ascent rule adaptively adjusts the penalty strength: $\lambda_{p,i}$ will increase whenever the corresponding constraint is violated (e.g., $\mathrm{J}(\hat{y}, \hat{y}_{p}) > \epsilon_{p,1}$), and decreases or leaves it unchanged when the constraint is satisfied. 


\begin{table*}[htbp]
\centering
\caption{F1 score (in \%) across 11 tasks. Column best is \textbf{bold} with dark-green background; runner-up is light-green.}
\label{tab:main_result}
\resizebox{\textwidth}{!}{
\begin{tabular}{l|ccccccccccc|c}
\toprule
\multirow{2}{*}{Model}
& \multicolumn{2}{c}{AggreFact}
& \multicolumn{2}{c}{TofuEval}
& \multirow{2}{*}{WiCE}
& \multirow{2}{*}{REVEAL}
& \multicolumn{1}{c}{Claim}
& \multicolumn{1}{c}{Fact}
& \multicolumn{1}{c}{Expert}
& \multirow{2}{*}{LFQA}
& \multicolumn{1}{c|}{RAG}
& \multirow{2}{*}{AVG}
\\
& CNN & XSum & MediaS & MeetB &  &  & Verify & Check & QA &  & Truth &  \\
\midrule
FactCC        & 78.30 & 69.20 & 82.46 & 86.74 & \ru{69.90} & 77.77 & 86.41 & 61.28 & 65.15 & 87.47 & 86.10 & 77.34 \\
SummaC-ZS       & 55.19 & 66.55 & 79.57 & 80.14 & 41.03 & 74.50 & 72.96 & 60.79 & 65.30 & 81.58 & 62.81 & 67.31 \\
SummaC-CV       & 66.41 & 48.56 & 69.70 & 77.87 & 50.98 & 70.77 & 68.36 & 57.53 & 65.21 & 79.16 & 56.92 & 64.68 \\
AlignScore      & 81.71 & 68.51 & 80.59 & 85.18 & 54.75 & 69.79 & 81.16 & 61.05 & 70.63 & 85.81 & 77.35 & 74.23 \\
MiniCheck-DBT   & 74.88 & 74.49 & 83.72 & 81.26 & 62.10 & 72.31 & 85.54 & 60.15 & 72.22 & 88.15 & 86.36 & 76.47 \\
MiniCheck-FT5 & 72.64 & 71.22 & 76.20 & \win{89.49} & 65.97 & 74.16 & 81.88 & \ru{63.39} & \ru{71.99} & 89.06 & 84.28 & 76.39 \\
FactCG-DBT      & 78.44 & 74.46 & \ru{83.98} & 80.07 & 69.87 & \win{80.53} & 83.45 & \win{63.42} & 58.54 & \ru{89.12} & 86.63 & 77.14 \\
FactCG-FT5    & 81.42 & \ru{74.52} & 82.36 & \ru{88.74} & 68.18 & 75.05 & \win{87.05} & 59.57 & 70.73 & 88.70 & \ru{86.82} & 78.47 \\
\midrule
CCHD-DBT       & \ru{88.05} & \win{77.10} & \win{84.74} & 87.43 & \win{70.70} & \ru{78.29} & 83.81 & 61.47 & 68.53 & \win{89.46} & \win{87.40} & \win{79.73} \\
CCHD-FT5 & \win{88.58} & 73.37 & 78.92 & 88.34 & 68.31 & 76.62 & \ru{86.50} & 62.46 & \win{72.41} & 88.28 & 86.42 & \ru{79.11} \\
\bottomrule
\end{tabular}
}
\end{table*}

\subsection{Instantiation of CCHD}
\label{sec:back_translation}


\textbf{Choice of paraphrase.}
Among common paraphrasing strategies (e.g., perturb \cite{agarwal2023perturb} and rewrite \cite{yadav2024pag}), we adopt \emph{back-translation} \cite{kryscinski2019factcc} for two practical reasons:
(i) it induces broad stylistic and lexical variation while largely preserving semantics; and 
(ii) it can be implemented reliably and inexpensively with off-the-shelf machine translation (MT) systems \cite{stahlberg2020translation_review}, e.g., a commercial MT API\footnote{\url{https://translate.google.com}}, thereby avoiding LLM-based paraphrasers and additional annotation.



\textbf{Extensions.}
CCHD naturally extends to multiple pivots and paraphrase strategies, at the cost of adding a few scalar dual multipliers $\lambda$ only. 
Cross-view consistency can also be enforced via pairwise constraints, e.g.,
$d_{\text{pred}}(\hat{y}_{p},\hat{y}_{q}) - \epsilon_{p,q} \le 0$ for two paraphrasers $p\neq q$,
which plug into the same Lagrangian without altering the optimization procedure.
In this paper, we adopt a \emph{single} pivot to control cost and compare various pivots in Sec. \ref{sec:sensitivity_analysis}.

\textbf{Efficiency.}
CCHD is efficient during both training and inference.
\emph{Training:} it is label-efficient (no additional human annotation) and parameter-efficient (introducing only per-view dual scalars).
\emph{Inference:} paraphrase generation is performed offline for training only, so test-time cost matches that of the underlying Flan-T5/DeBERTa detector, preserving the runtime profile of standard discriminative baselines~\cite{tang2024minicheck,lei2025factcg}.

\section{Experiments}
\label{sec:experiments}

\subsection{Experiment Settings}
\label{sec:experiemnt_settings}

\textbf{Model setup.}
We instantiate the detector with \textit{DeBERTa}~\cite{he2020deberta} (default) or \textit{Flan-T5}~\cite{chung2022flant5} without architectural changes, denoted with suffix -DBT and -FT5, respectively.
A single pivot language (French by default) is used for back-translation to construct one paraphrase view per example.
Analyses of backbone choice, pivot languages, and constraint design appear in Secs.~\ref{sec:ablation} and \ref{sec:sensitivity_analysis}.

\textbf{Hyperparameters.}
The major settings follow FactCG~\cite{lei2025factcg}. 
The primal learning rate $\eta_{0}$ is $1\!\times\!10^{-5}$ for DeBERTa and $5\!\times\!10^{-5}$ for Flan-T5.
All dual variables use $\eta_{p,i}=1\!\times\!10^{-3}$.
We set all tolerances $\epsilon_{p,i}=0$ and initialize the Lagrange multipliers $\boldsymbol{\lambda}$ from a uniform distribution.

\textbf{Baselines.}
We compare against state-of-the-art non-LLM detectors: \textit{FactCC}~\cite{kryscinski2019factcc}, \textit{SummaC}~\cite{laban2022summac}, \textit{AlignScore}~\cite{zha2023alignscore}, \textit{MiniCheck}~\cite{tang2024minicheck}, and \textit{FactCG}~\cite{lei2025factcg}. 
We use the authors' released implementations and recommended checkpoints when available. 
Besides, we calibrate a decision threshold on the development split following each paper's protocol.

\textbf{Datasets and metrics.}
Evaluation follows the \textbf{LLM-AggreFact} benchmark, which aggregates 11 factuality/grounded-consistency datasets~\cite{tang2024minicheck}. 
We adopt the official task definitions and splits, and report per-task F1 as well as macro-averaged F1 scores (AVG).

\subsection{Main Results}
\label{sec:main_results}

Table~\ref{tab:main_result} reports F1 (\%) across the 11 tasks in LLM-AggreFact.
CCHD-DBT and CCHD-FT5 achieve the strongest average performance, with macro-F1 of \textbf{79.73} and \textbf{79.11}, respectively.
CCHD-DBT exceeds the best non-CCHD baseline (FactCG-FT5, 78.47) by \textbf{+1.26} points while retaining the efficiency of discriminative detectors (with $\sim$50\% fewer parameters).
Regarding per-task performance,
CCHD is the winner on the majority of tasks (7/11), indicating robustness and generalization across tasks. 

\textbf{Across backbones.}
Both DeBERTa and Flan-T5 instantiations of CCHD outperform their backbone-matched baselines, suggesting that the gains arise from the constrained objective with GDA rather than backbone-specific factors.

\textbf{Summary.}
Optimizing paraphrase-consistency and label-preserving constraints via GDA yields consistent, cross-dataset improvements without inference-time overhead, establishing CCHD as a superior discriminative detector.

\begin{table*}[ht]
\centering
\caption{CCHD variants adopt \texttt{fr}, \texttt{zh}, and \texttt{es} as pivot languages; and enforce embedding- or prediction-level constraints.}
\label{tab:sensitivity}
\resizebox{\textwidth}{!}{
\begin{tabular}{l|ccccccccccc|c}
\toprule
\multirow{2}{*}{Model}
& \multicolumn{2}{c}{AggreFact}
& \multicolumn{2}{c}{TofuEval}
& \multirow{2}{*}{WiCE}
& \multirow{2}{*}{REVEAL}
& \multicolumn{1}{c}{Claim}
& \multicolumn{1}{c}{Fact}
& \multicolumn{1}{c}{Expert}
& \multirow{2}{*}{LFQA}
& \multicolumn{1}{c|}{RAG}
& \multirow{2}{*}{AVG}
\\
& CNN & XSum & MediaS & MeetB &  &  & Verify & Check & QA &  & Truth &  \\
\midrule
\texttt{fr}-embd & 87.24 & 74.13 & \ru{84.58} & 82.00 & 71.13 & \ru{79.45} & \ru{84.44} & 57.82 & 62.60 & 89.26 & \win{88.24} & 78.26 \\
\texttt{fr}-pred & \ru{88.05} & \win{77.10} & \win{84.74} & \ru{87.43} & 70.70 & 78.29 & 83.81 & 61.47 & \win{68.53} & 89.46 & 87.40 & \win{79.73} \\
\midrule
\texttt{zh}-embd & 75.42 & \ru{75.32} & 83.40 & 81.74 & \win{77.18} & 76.94 & \win{84.64} & \win{69.94} & 64.32 & 83.47 & 85.92 & 78.03 \\
\texttt{zh}-pred & \win{90.11} & 73.12 & 83.99 & \win{87.49} & 68.84 & 77.05 & 83.88 & \ru{69.27} & 59.12 & 89.87 & 84.78 & \ru{78.86} \\
\midrule
\texttt{es}-embd & 84.03 & 71.94 & 81.25 & 80.62 & \ru{72.96} & 77.50 & 83.99 & 63.71 & \ru{65.74} & \ru{89.97} & \ru{87.92} & 78.15 \\
\texttt{es}-pred & 82.84 & 74.35 & 79.84 & 83.29 & 70.69 & \win{81.66} & 82.68 & 65.77 & 63.27 & \win{90.13} & 87.48 & 78.36 \\
\bottomrule

\end{tabular}
}
\end{table*}

\subsection{Ablation Study}
\label{sec:ablation}

To disentangle the effects of the GDA scheme and the two constraint families in Eqs.~\eqref{eq:c1} and \eqref{eq:c2}, we evaluate:
\emph{C1 only} (paraphrase-consistency, Eq.~\eqref{eq:c1}),
\emph{C2 only} (label-preservation, Eq.~\ref{eq:c2}),
and \emph{Fixed} $\lambda$ (include both constraints with fixed $\lambda$; no GDA).

\begin{figure}[tbp]
    \centering
    \includegraphics[width=0.95\linewidth]{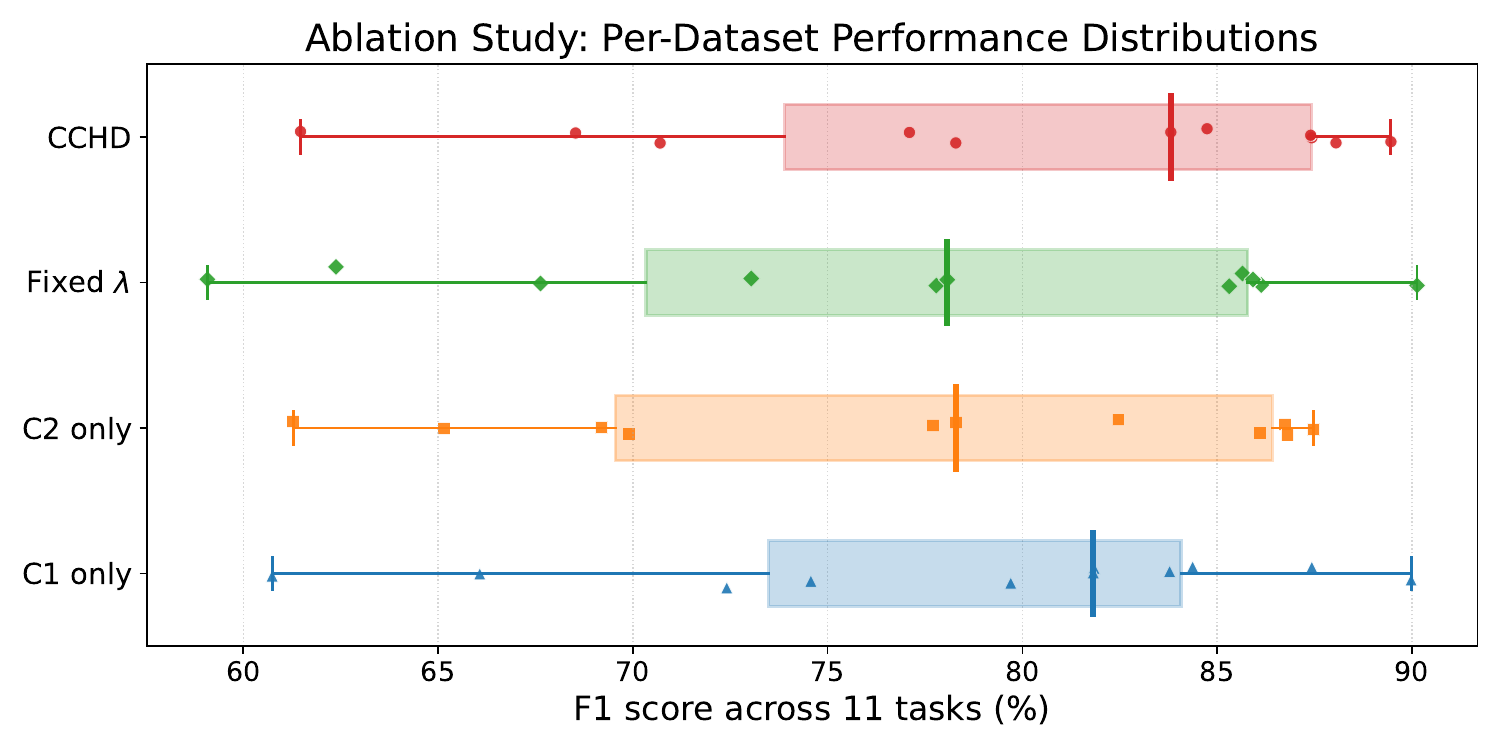}
    \caption{\textbf{Ablation studies.} 
    Per-task F1 boxplots over 11 factuality tasks for CCHD and three variants: 
    \textit{C1 only} and \textit{C2 only} optimizes the constraint \eqref{eq:c1} and \eqref{eq:c2}, respectively. 
    \textit{Fixed} $\lambda$ considers both constraints without GDA strategy.}
    \label{fig:ablation_study}
    \vspace{-.5cm}
\end{figure}

\textbf{Overall effect of GDA and joint constraints.}
As visualized in Fig.~\ref{fig:ablation_study}, the per-task F1 distribution of CCHD is noticeably right-shifted relative to all ablated variants. 
Macro-averaged over 11 tasks, CCHD attains the highest F1 of \textbf{79.73}, exceeding \emph{C1 only} (78.43; +1.30\,pp), \emph{C2 only} (77.38; +2.36\,pp), and \emph{Fixed} $\lambda$ (77.29; +2.44\,pp).

\textbf{Effect of each constraint.}
\emph{C1 only} leverages paraphrase-consistency and attains the second-best macro-F1, but can drift under noisy paraphrases without an explicit label tether. 
\emph{C2 only} anchors paraphrases to ground-truth labels, yet is sensitive to augmentation noise and less invariant to paraphrastic variation. 
\textit{CCHD} jointly enforces both constraints with \textit{adaptive} multipliers, and strikes the best accuracy across tasks.

\subsection{Sensitivity Analysis}
\label{sec:sensitivity_analysis}

\textbf{Setup.} 
Using the DeBERTa backbone, 
we vary (i) the \emph{pivot language} for back-translation: French (\texttt{fr}), Chinese (\texttt{zh}), Spanish (\texttt{es});
and (ii) the \emph{paraphrase-consistency design}: a \emph{prediction-level} constraint (``pred'', Eq.~\eqref{eq:c1}) that penalizes divergence between output distributions, 
and an \emph{embedding-level} constraint (``embd'') that penalizes the $\ell_{1}$ distance between [\texttt{CLS}] representations. 
Results appear in Table~\ref{tab:sensitivity}.

\textbf{Comparison to baselines.}
Across the six CCHD variants, macro-F1 ranges from \textbf{78.03} to \textbf{79.73}. 
Every variant \emph{significantly exceeds all DeBERTa-based baselines} in Table~\ref{tab:main_result} (FactCC, AlignScore, MiniCheck-DBT, FactCG-DBT). 

\textbf{Effect of pivot language.}
All three pivots are competitive, but French yields the strongest averages under both designs:
\texttt{fr}-pred and \texttt{fr}-embd achieves the best overall macro-F1 comparing to \texttt{zh} and \texttt{es} counterparts. 
However, task-level performance is complementary:
\texttt{zh}-pred attains the top scores on \emph{AggreFact-CNN} and \emph{TofuEval-MeetB}; while \texttt{es}-pred leads on \emph{REVEAL} and \emph{LFQA}. 
It could because \texttt{fr}$\leftrightarrow$\texttt{en} MT systems are historically strong and syntactically closer than \texttt{zh}$\leftrightarrow$\texttt{en} and \texttt{es}$\leftrightarrow$\texttt{en}. 
It may yield paraphrases that better preserve semantics while still varying surface form (less semantic drift), thus providing a cleaner consistency signal. 

\textbf{Effect of constraint design.}
Prediction-level consistency outperforms embedding-level consistency for each pivot.
While embd can be favorable on structure-sensitive tasks (e.g., \emph{WiCE}, \emph{Claim Verify}), pred dominates on most others and yields more stable gains overall.
The potential reason could be aligning \emph{predictions} directly regularizes the decision boundary and complements the primary cross-entropy, whereas embedding-level constraints are sensitive to representation scaling and may under- or over-constrain learning.


\section{Conclusion}

We presented the \emph{Consistency-Constrained Hallucination Detector}, which enfores paraphrase consistency and label-preservation constraints. 
Instantiated via back-translation, CCHD delivers significant F1 gains across 11 factuality datasets without extra inference time. 
%
To further enhance the generalization and reliability of CCHD, 
we will explore: 
(i) combining multiple pivot languages and non-MT paraphrasers with adaptive selection; 
and (ii) extending evaluation to hallucinations produced by \emph{multimodal} LLMs \cite{bai2024mllm_hallucination_review}. 

\section*{Acknowledgments}
This work was supported by the Natural Science Foundation of China (No. 62476060).
Chao Chen was also supported by the National Key Research and Development Program of China (No. 2023YFB3106504).

\newpage
\bibliographystyle{IEEEbib}
\bibliography{main}

@article{shakil2024summary_review,
  title={Abstractive text summarization: State of the art, challenges, and improvements},
  author={Shakil, Hassan and Farooq, Ahmad and Kalita, Jugal},
  journal={Neurocomputing},
  volume={603},
  pages={128255},
  year={2024},
  publisher={Elsevier}
}

@article{stahlberg2020translation_review,
  title={Neural machine translation: A review},
  author={Stahlberg, Felix},
  journal={Journal of Artificial Intelligence Research},
  volume={69},
  pages={343--418},
  year={2020}
}

@article{bai2024mllm_hallucination_review,
  title={Hallucination of multimodal large language models: A survey},
  author={Bai, Zechen and Wang, Pichao and Xiao, Tianjun and He, Tong and Han, Zongbo and Zhang, Zheng and Shou, Mike Zheng},
  journal={arXiv preprint arXiv:2404.18930},
  year={2024}
}

@article{lei2025factcg,
  title={FactCG: Enhancing fact checkers with graph-based multi-hop data},
  author={Lei, Deren and Li, Yaxi and Li, Siyao and Hu, Mengya and Xu, Rui and Archer, Ken and Wang, Mingyu and Ching, Emily and Deng, Alex},
  journal={arXiv preprint arXiv:2501.17144},
  year={2025}
}

@article{laban2022summac,
  title={SummaC: Re-visiting NLI-based models for inconsistency detection in summarization},
  author={Laban, Philippe and Schnabel, Tobias and Bennett, Paul N and Hearst, Marti A},
  journal={Transactions of the Association for Computational Linguistics},
  volume={10},
  pages={163--177},
  year={2022},
  publisher={MIT Press One Rogers Street, Cambridge, MA 02142-1209, USA journals-info~…}
}

@article{zha2023alignscore,
  title={AlignScore: Evaluating factual consistency with a unified alignment function},
  author={Zha, Yuheng and Yang, Yichi and Li, Ruichen and Hu, Zhiting},
  journal={arXiv preprint arXiv:2305.16739},
  year={2023}
}

@article{tang2024minicheck,
  title={Minicheck: Efficient fact-checking of llms on grounding documents},
  author={Tang, Liyan and Laban, Philippe and Durrett, Greg},
  journal={arXiv preprint arXiv:2404.10774},
  year={2024}
}

@article{manakul2023selfcheckgpt,
  title={Selfcheckgpt: Zero-resource black-box hallucination detection for generative large language models},
  author={Manakul, Potsawee and Liusie, Adian and Gales, Mark JF},
  journal={arXiv preprint arXiv:2303.08896},
  year={2023}
}

@article{muhammed2025selfcheckagent,
  title={SelfCheckAgent: Zero-Resource Hallucination Detection in Generative Large Language Models},
  author={Muhammed, Diyana and Rabby, Gollam and Auer, S{\"o}ren},
  journal={arXiv preprint arXiv:2502.01812},
  year={2025}
}

@article{kryscinski2019factcc,
  title={Evaluating the factual consistency of abstractive text summarization},
  author={Kry{\'s}ci{\'n}ski, Wojciech and McCann, Bryan and Xiong, Caiming and Socher, Richard},
  journal={arXiv preprint arXiv:1910.12840},
  year={2019}
}

@inproceedings{lin2020gda,
  title={On gradient descent ascent for nonconvex-concave minimax problems},
  author={Lin, Tianyi and Jin, Chi and Jordan, Michael},
  booktitle={International conference on machine learning},
  pages={6083--6093},
  year={2020},
  organization={PMLR}
}

@article{yang2025metaqa,
  title={Hallucination Detection in Large Language Models with Metamorphic Relations},
  author={Yang, Borui and Al Mamun, Md Afif and Zhang, Jie M and Uddin, Gias},
  journal={Proceedings of the ACM on Software Engineering},
  volume={2},
  number={FSE},
  pages={425--445},
  year={2025},
  publisher={ACM New York, NY, USA}
}

@inproceedings{bousquet2023jeffreys,
  title={Jeffreys divergence-based regularization of neural network output distribution applied to speaker recognition},
  author={Bousquet, Pierre-Michel and Rouvier, Mickael},
  booktitle={ICASSP 2023-2023 IEEE International Conference on Acoustics, Speech and Signal Processing (ICASSP)},
  pages={1--5},
  year={2023},
  organization={IEEE}
}

@article{chung2022flant5,
  title = {Scaling Instruction-Finetuned Language Models},
  author = {Chung, Hyung Won and Hou, Le and Longpre, Shayne and others},
  journal = {arXiv:2210.11416},
  year = {2022}
}

@article{he2020deberta,
  title={Deberta: Decoding-enhanced bert with disentangled attention},
  author={He, Pengcheng and Liu, Xiaodong and Gao, Jianfeng and Chen, Weizhu},
  journal={arXiv preprint arXiv:2006.03654},
  year={2020}
}

@article{maynez2020faithfulness,
  title={On faithfulness and factuality in abstractive summarization},
  author={Maynez, Joshua and Narayan, Shashi and Bohnet, Bernd and McDonald, Ryan},
  journal={arXiv preprint arXiv:2005.00661},
  year={2020}
}

@inproceedings{chen2021snx,
  title={Self-learn to explain siamese networks robustly},
  author={Chen, Chao and Shen, Yifan and Ma, Guixiang and Kong, Xiangnan and Rangarajan, Srinivas and Zhang, Xi and Xie, Sihong},
  booktitle={2021 IEEE International Conference on Data Mining (ICDM)},
  pages={1018--1023},
  year={2021},
  organization={IEEE}
}

@inproceedings{laban2023summedits,
  title={SummEdits: Measuring LLM ability at factual reasoning through the lens of summarization},
  author={Laban, Philippe and Kry{\'s}ci{\'n}ski, Wojciech and Agarwal, Divyansh and Fabbri, Alexander Richard and Xiong, Caiming and Joty, Shafiq and Wu, Chien-Sheng},
  booktitle={Proceedings of the 2023 conference on empirical methods in natural language processing},
  pages={9662--9676},
  year={2023}
}

@inproceedings{yadav2024pag,
  title={Pag-llm: Paraphrase and aggregate with large language models for minimizing intent classification errors},
  author={Yadav, Vikas and Tang, Zheng and Srinivasan, Vijay},
  booktitle={Proceedings of the 47th international ACM SIGIR conference on research and development in information retrieval},
  pages={2569--2573},
  year={2024}
}

@inproceedings{agarwal2023perturb,
  title={Towards effective paraphrasing for information disguise},
  author={Agarwal, Anmol and Gupta, Shrey and Bonagiri, Vamshi and Gaur, Manas and Reagle, Joseph and Kumaraguru, Ponnurangam},
  booktitle={European Conference on Information Retrieval},
  pages={331--340},
  year={2023},
  organization={Springer}
}

\end{document}